# Status of the XTAG System

C. Doran, D. Egedi, B. A. Hockey, B. Srinivas
Institute for Research in Cognitive Science
University of Pennsylvania
Philadelphia, PA 19104-6228, USA
{cdoran, egedi, beth, srini}@unagi.cis.upenn.edu

**Abstract**
XTAG is an ongoing project to develop a wide-coverage grammar for English, based on the Feature-based Lexicalized Tree Adjoining Grammar (FB-LTAG) formalism. The XTAG system integrates a morphological analyzer, an N-best part-of-speech tagger, an Earley-style parser and an X-window interface, along with a wide-coverage grammar for English developed using the system. This system serves as a linguist's workbench for developing FB-LTAG specifications. This paper presents a description of and recent improvements to the various components of the XTAG system. It also presents the recent performance of the wide-coverage grammar on various corpora and compares it against the performance of other wide-coverage and domain-specific grammars.

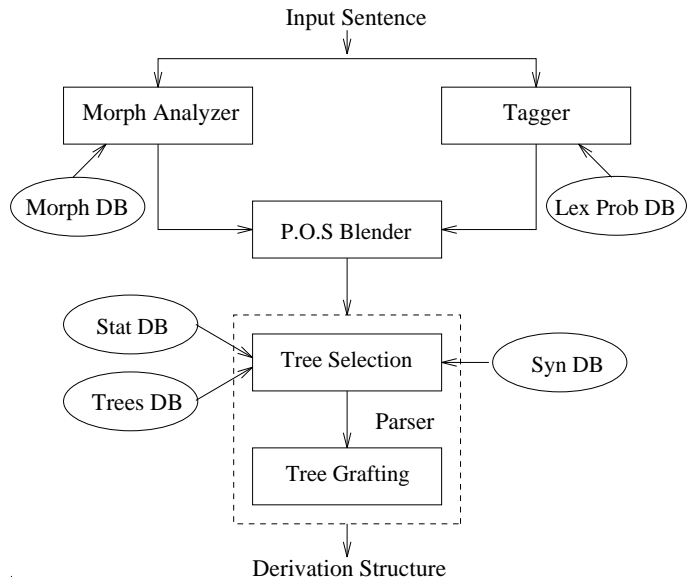

Figure 1: Overview of XTAG system

## 1 Introduction

XTAG is an on-going project to develop a wide-coverage grammar for English, based on the Feature-Based Lexicalized Tree Adjoining Grammar (FB-LTAG) formalism.[1] FB-LTAG is a lexicalized mildly-context sensitive tree rewriting system [5, 6] that is closely related to Dependency Grammars and Categorial Grammars. The XTAG system serves as a workbench for the development of FB-LTAGs. XTAG consists of a predictive left-to-right parser, an X-window interface, a morphological analyzer, and a part-of-speech tagger (also referred to as simply 'tagger') along with a wide-coverage grammar for English.

## 2 System Description

Figure 1 shows the overall flow of the system when parsing a sentence. The input sentence is submitted to the **Morphological Analyzer** and the **Tagger**. The morphological analyzer retrieves the morphological information for each word from the morphological database. This output is filtered in the **P.O.S Blender** using the output of the trigram tagger to reduce the part-of-speech ambiguity of the words. The sentence, now annotated with part-of-speech tags and morphological information for each word, is input to the **Parser**, which consults the syntactic database and tree database to retrieve the appropriate tree structures for each lexical item. A variety of heuristics are used to reduce the number of trees selected. The parser then combines the structures to obtain the parse(s) of the sentence.

A summary of each component is presented in Table 1.

## 3 English Grammar

The morphological, syntactic, and tree databases together comprise the English grammar. Lexical items not in the databases are handled by default mechanisms. The range of syntactic phenomena that can be handled is large and includes auxiliaries (including inversion), copula, raising and small clause constructions, topicalization, relative clauses, infinitives, gerunds, passives, adjuncts, it-clefts, wh-clefts, PRO constructions, noun-noun modifications, extraposition, determiner phrases, genitives, negation, noun-verb contractions, sentenital adjuncts and imperatives. The combination of a large lexicon and wide phenomena coverage result in a robust system. The XTAG grammar has been relatively stable since November, 1993, although new analyses are still being added periodically. Analyses of time NPs and bare infinitives are currently under development.

---
[1] XTAG is available via ftp. Instructions and more information can be obtained by mailing requests to xtag-request@linc.cis.upenn.edu.

| Component | Details |
|---|---|
| Morphological Analyzer and Morph Database | Consists of approximately 317,000 inflected items. Entries are indexed on the inflected form and return the root form, POS, and inflectional information. Database does not address derivational morphology. |
| POS Tagger and Lex Prob Database | Wall Street Journal-trained trigram tagger [3] extended to output N-best POS sequences [7]. Decreases the time to parse a sentence by an average of 93%. |
| Syntactic Database | More than 105,000 entries. Each entry consists of: the uninflected form of the word, its POS, the list of trees or tree-families associated with the word, and a list of feature equations that capture lexical idiosyncrasies. |
| Tree Database | 566 trees, divided into 40 tree families and 62 individual trees. Tree families represent subcategorization frames; the trees in a tree family would be related to each other transformationally in a movement-based approach. |
| X-Interface | Menu-based facility for creating and modifying tree files. User controlled parser parameters: parser's start category, enable/disable/retry on failure for POS tagger. Storage/retrieval facilities for elementary and parsed trees as text and postscript files. Graphical displays of tree and feature data structures. Hand combination of trees by adjunction or substitution for diagnosing grammar problems. |

Table 1: System Summary

## 4 Recent Developments

Development of database maintenance tools, parsing and evaluation of the coverage of the system on various corpora have been some of the recent developments on the XTAG project.

### 4.1 Database Maintenance Tool Development

The morphological and the syntactic information is available in both the ASCII format as well as an binary-encoded database format. The ASCII format is well-suited for various UNIX utilities while the database format is used for fast access during program execution. However even the ASCII formatted representation is not well-suited for human readability. An X-windows interface[2] for the syntactic database allows users to easily look at the database. Searching for specific information on certain fields of the database are also available. Also, the interface allows a user to insert, delete and update any information in the database. Figure 2 shows the interface for the morphology database and Figure 3 shows the interface for the syntactic database.

### 4.2 Parsing Corpora

The natural step after developing the sizeable grammar is to evaluate and compare XTAG's performance against other grammar systems. XTAG has been used to parse sentences from the Wall-Street Journal (WSJ), the IBM manual, and the ATIS corpus. The XTAG parsed corpus consists of all the derivations obtained for each sentence for which the system found a parse. These derivations have been used to evaluate and improve the performance of the system in the ways discussed below.

---
[2]The interface uses the MIT Athena Toolkit, which is distributed with the standard MIT X release.

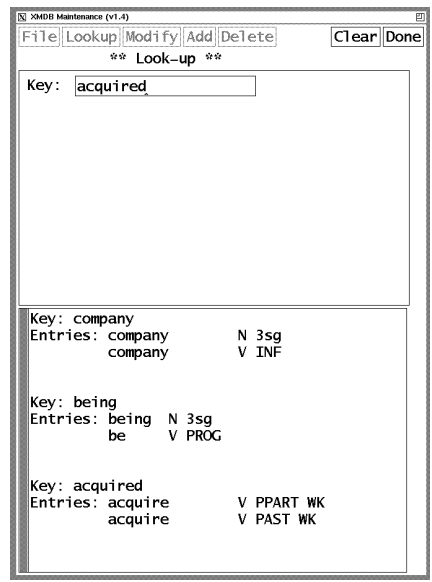

Figure 2: Interface to the Morphology database

### 4.3 Statistics Database

The statistics database contains tree unigram frequencies which have been collected from the XTAG-parsed corpus. The parser, using information from the statistics database, assigns each word of the input sentence the top three most frequently used trees given the part-of-speech of the word. On failure, the parser retries using all the trees suggested by the syntactic database for each word. The augmented parser has been observed to have a success rate of 50% without retries. Due to the sparseness of data, the corpus unigram information is currently over POS tag/tree pairs. We will ultimately have statistics for lexical item/tree pairs after parsing more sentences.

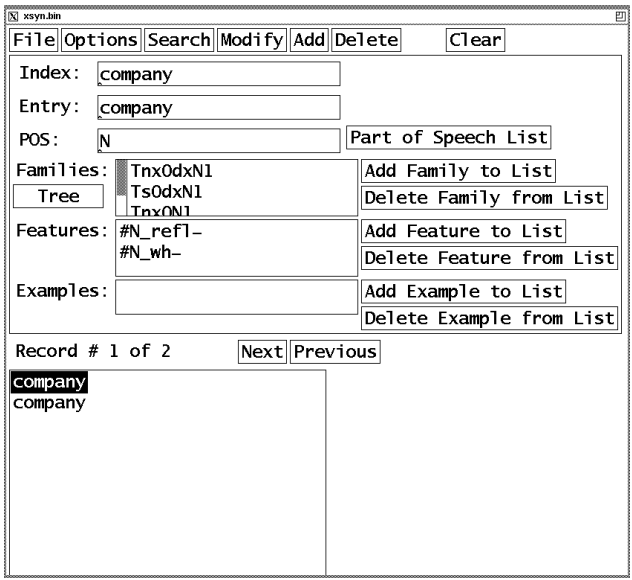

Figure 3: Interface to the Syntactic database

### 4.4 Performance Evaluation

Table 2 contains the preliminary results from evaluating the coverage and correctness of the XTAG system on the WSJ, IBM manual, and ATIS corpora. For this evaluation, a sentence is considered to have parsed correctly if XTAG produces parse trees. Verifying the presence of the correct parse among the parses generated is done manually at present. Sentence fragments are not included in the data below; XTAG is currently being extended to handle sentence fragments. The performance results do not involve any tuning or training on any particular corpora.

| Corpus | # of Sentences | % Parsed | Av. # of parses/sent |
|---|---|---|---|
| WSJ | 18730 | 41.22% | 7.46 |
| IBM Manual | 2040 | 75.42% | 6.12 |
| ATIS | 524 | 88.35% | 6.00 |

Table 2: Performance on various corpora

A more detailed experiment to measure the crossing bracket accuracy of the XTAG-parsed IBM-manual sentences has been performed. XTAG-parses of 1100 IBM-manual sentences have been compared[3] against the bracketing given in the Lancaster treebank of IBM-manual sentences[4].

Table 3 shows the results obtained in this experiment. It also shows the crossing bracket accuracy of the latest IBM statistical parser [4] and its recall and precision on the same genre of sentences. Recall is defined as a measure of the number of bracketed constituents the system got right divided by the number of constituents in the corresponding Treebank sentences. Precision is defined as the number of bracketed constituents the system got right divided by the number of bracketed constituents in the system's parse.

---
[3] We used the parseval program written by Phil Harison (phil@atc.boeing.com).
[4] The treebank was obtained through Salim Roukos (roukos@watson.ibm.com) at IBM.

| System | # of sentences | Crossing Brackets | Recall | Precision |
|---|---|---|---|---|
| XTAG | 1100 | 80% | 84.32% | 59.28% |
| IBM Stat. parser | 1100 | 86.2% | 86.20% | 85.00% |

Table 3: Performance on IBM-manual sentences

The reason for the misleadingly low precision is due to the fact that the XTAG parse is much more detailed in terms of constituent structure when compared to that of the Lancaster treebank parses which provide a very skeletal representation of phrases. Table 4 illustrates this quantitatively in terms of the number of constituents produced by XTAG and IBM-parser for sentences of varying lengths.

| System | Sent. Length | # of sent | Av. # of words/sent | Av. # of Consts/sent |
|---|---|---|---|---|
| XTAG | 1-10 | 654 | 7.45 | 22.03 |
|  | 1-15 | 978 | 9.13 | 30.56 |
| IBM Stat. Grammar | 1-10 | 447 | 7.50 | 4.60 |
|  | 1-15 | 883 | 10.30 | 6.40 |

Table 4: Distribution of sentences, words/sentence and constituents/sentence

We compared the XTAG system to the Alvey Natural Language Tools (ANLT) Parser, and found that the two performed comparably. We parsed the same set of 143 LDOCE Noun Phrases as presented in Appendix B of the technical report [2] using the XTAG parser. We also compared the total number of derivations obtained from XTAG with that obtained from the ANLT parser. Table 5 summarizes the results of this experiment. For the XTAG system, performance results with and without the POS tagger are shown.

| System | # parsed | % parsed | Max dervs | Av dervs |
|---|---|---|---|---|
| ANLT Parser | 127 | 88.81% | 32 | 4.57 |
| XTAG Parser with tagger | 93 | 65.03% | 28 | 3.45 |
| XTAG Parser w/o tagger | 124 | 86.71% | 28 | 4.14 |

Table 5: Comparison of XTAG and ANLT Parser

We also compared XTAG system against the CLARE parser [1] and found that the two performed comparably. Table 6 summarizes the performance of XTAG and CLARE-2 system[5].

---
[5] It is estimated that the performance of CLARE-3 system is about 10% better than the performance of CLARE-2, in general

| System | Mean length | % parsed |
|---|---|---|
| CLARE-2 | 6.53 | 68.50% |
| XTAG | 7.62 | 88.35% |

Table 6: Performance of XTAG and CLARE-2 on the ATIS domain

In order to contrast the performance of XTAG on a corpus such as Wall Street Journal that has more structural variations than the sentence that appear in ATIS we compared the performance of XTAG against the performance of CLARE-2 on LOB corpus. Table 7 shows the results of this comparison.

| System | Corpus | Mean Length | % parsed |
|---|---|---|---|
| CLARE-2 | LOB | 5.95 | 53.40% |
| XTAG | WSJ | 6.00 | 55.58% |

Table 7: Performance of CLARE-2 on LOB and XTAG on WSJ corpora